\documentclass[10pt]{article} 

\usepackage[accepted]{rlj}           

\usepackage{amssymb}            
\usepackage{mathtools}          
\usepackage{mathrsfs}           
\usepackage{graphicx}           
\usepackage{subcaption}         
\usepackage[space]{grffile}     
\usepackage{url}                
\usepackage{lipsum}             
\usepackage{titletoc}
\usepackage{amsthm}
\usepackage{amsmath}
\usepackage{booktabs}
\usepackage{multirow}
\usepackage{wrapfig}
\usepackage[font=small]{caption}
\usepackage[hidelinks, colorlinks=true, citecolor=lightgray, linkcolor=lightgray]{hyperref}
\definecolor{memory-green}{HTML}{3B7D23}
\definecolor{action-purple}{HTML}{78206E}
\definecolor{dynamics-misid-blue}{HTML}{16639c}
\definecolor{rewards-misid-orange}{HTML}{c33800}
\definecolor{both-misid-green}{HTML}{168336}
\definecolor{obs-blue}{HTML}{002060}

\DeclareMathOperator*{\argminA}{arg\,min} 

\title{Zero-Shot Reinforcement Learning \\ Under Partial Observability}

\setrunningtitle{Zero-Shot Reinforcement Learning Under Partial Observability}

\author{Scott Jeen\textsuperscript{1}, Tom Bewley\textsuperscript{2}, Jonathan M. Cullen\textsuperscript{1}}

\emails{srj38@cam.ac.uk}

\affiliations{
$^{1}$\textbf{University of Cambridge}\\
$^{2}$\textbf{Independent}\\
\par 
}

\contribution{
    We explore the empirical failure modes of state-of-the-art zero-shot RL methods (specifically forward-backward representations, or FB) given partially observed (noisy) states.
    }
    {
    None
    }

\contribution{
    We present a new architecture called FB with memory (FB-M) which has a memory-based forward model $F$, backward model $B$ and policy $\pi$. Though we develop our method within the FB framework, our proposals are fully compatible with
    other
    zero-shot RL methods.  
    }
    {
    Prior zero-shot RL methods, including FB \citep{touati2021learning} and USF-based HILP \citep{borsa2018, park2024foundation}, are memory-free.   
    }

\contribution{
    We show that, in aggregate, FB-M outperforms memory-free FB and HILP, as well as a na\"ive observation-stacking baseline, in domains where the states are noisy or randomly dropped, or where there is a change in dynamics function between training and testing.   
    }
    {
    None
    }

\contribution{
    We report better performance when the memory model is a GRU than when it is a transformer or S4d model.
    }
    {
    This aligns with \cite{morad2023popgym}'s finding that GRUs were the most performant memory model on POPGym, a partially observed single-task RL benchmark.
    }


\summary{Recent work has shown that, under certain assumptions, zero-shot reinforcement learning (RL) methods can generalise to \textit{any} unseen task in an environment after reward-free pre-training. Access to Markov states is one such assumption, yet, in many practical applications, the Markov state is only \textit{partially observed} via unreliable or incomplete observations. Here, we explore how the performance of standard zero-shot RL methods degrades when subjected to partially observability, and show that, as in single-task RL, memory-based architectures are an effective remedy. We evaluate our \textit{memory-based} zero-shot RL methods in domains where we simulate unreliable states by adding noise or dropping them randomly, and in domains where we simulate incomplete observations by changing the dynamics between training and testing rewards without communicating the change to the agent. In these settings, our proposals show improved performance over memory-free baselines, which we pay for with slower, less stable training dynamics.
}

\begin{document}

\maketitle  

\begin{abstract}
Recent work has shown that, under certain assumptions, zero-shot reinforcement learning (RL) methods can generalise to \textit{any} unseen task in an environment after reward-free pre-training. Access to Markov states is one such assumption, yet, in many real-world applications, the Markov state is only \textit{partially observable}. Here, we explore how the performance of standard zero-shot RL methods degrades when subjected to partially observability, and show that, as in single-task RL, memory-based architectures are an effective remedy. We evaluate our \textit{memory-based} zero-shot RL methods in domains where the states, rewards and a change in dynamics are partially observed, and show improved performance over memory-free baselines. Our code is open-sourced via the project page: \url{https://enjeeneer.io/projects/bfms-with-memory/}.
\end{abstract}

\section{Introduction}
Large-scale unsupervised pre-training has proven an effective recipe for producing vision \citep{rombach2022high} and language \citep{brown2020language} models that generalise to unseen tasks. The zero-shot reinforcement learning (RL) problem setting \citep{touati2022does} requires us to produce sequential decision-making agents with similar generality. It asks, informally: can we pre-train agents from datasets of reward-free trajectories such that they can immediately generalise to \textit{any} unseen reward function at test time? A family of methods called \textit{behaviour foundation models} (BFMs) \citep{touati2021learning, jeen2024zeroshot, pirotta2024fast} theoretically solve the zero-shot RL problem under certain assumptions \citep{touati2021learning}, and empirically return near-optimal policies for many unseen goal-reaching and locomotion tasks \citep{touati2022does}.

These results have assumed access to Markov states that provide all the information the agent requires to solve a task. Though this is a common assumption in RL, for many interesting problems, the Markov state is only \textit{partially observed} via unreliable or incomplete observations \citep{kaelbling1998}. Observations can be unreliable because of sensor noise or issues with telemetry \citep{meng2021memory}. Observations can be incomplete because of egocentricity \citep{tirumala2024learning}, occlusions \citep{heess2015memory} or because they do not communicate a change to the environment's task or dynamics context \citep{hallak2015contextual}.     

\begin{figure}[t]
\centering
    \includegraphics[width=\textwidth]{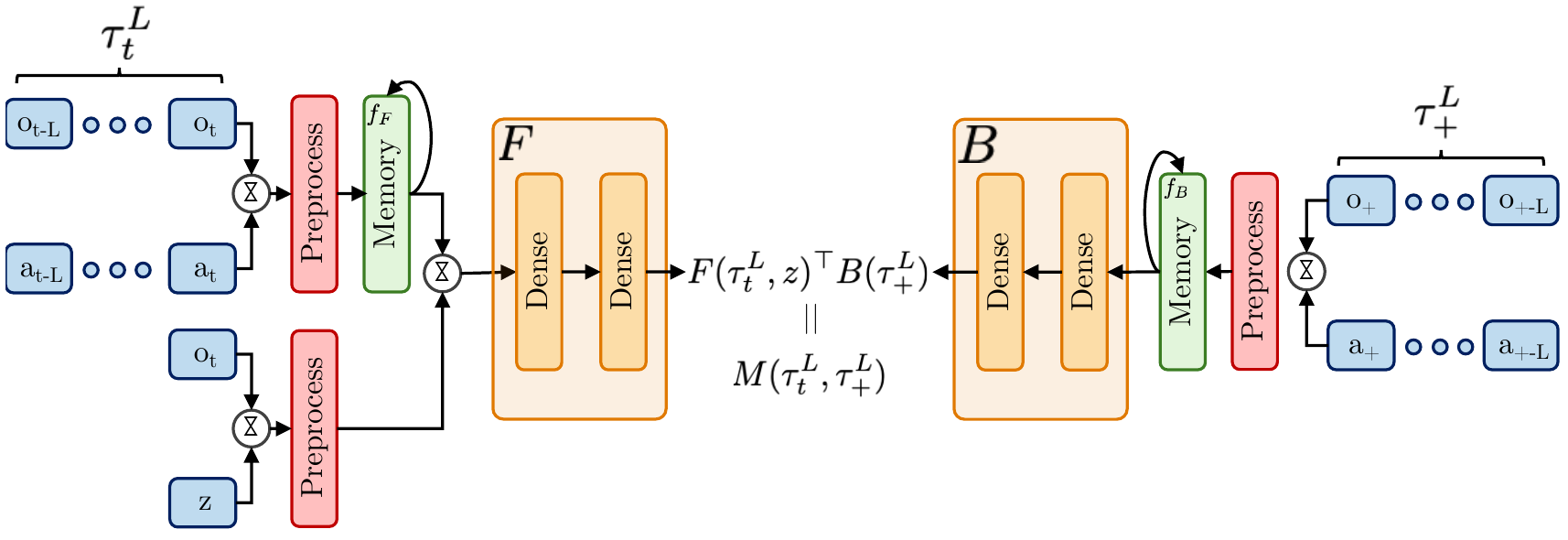}
    \caption{\textbf{BFMs with memory}.
    In the case of FB, the forward model $F$ and backward model $B$ condition on the output of memory models that compress trajectories of observations and actions. According to standard FB theory, their dot product predicts $M^{\pi_z}(\tau_t^L, \tau_+^L)$, the successor measure from $L$-length trajectory $\tau_t^L$ to $L$-length future trajectory $\tau_+^L$, from which a $Q$ function can be derived. Figure \ref{fig: fb architecture} in Appendix \ref{appendix: implementation details} illustrates memory-free FB for comparison.}
    \label{fig: confb architecture}
\end{figure}

How do BFMs fare when subjected to partial observability? That is the primary question this paper seeks to answer, and one we address in three parts. First, we expose the mechanisms that cause the performance of standard BFMs to degrade under partial observability (Section \ref{subsection: failure mode of existing methods}). Second, we repurpose methods that handle partial observability in single-task RL for use in the zero-shot RL setting, that is, we add \textit{memory models} to the BFM framework (Section \ref{subsection: method}, Figure \ref{fig: confb architecture}). Third, we conduct experiments that test how well BFMs augmented with memory models manage partially observed states (Section \ref{subsection: partially observed states}) and partially observed changes in dynamics (Section \ref{subsection: partially observed dynamics}). We conclude by discussing limitations and next steps.

\section{Related Work}\label{section: related work}

\vspace{-0.1cm}
\subsection{Zero-shot RL}
\vspace{-0.1cm}

\textbf{Offline RL} \; An important part of the zero-shot RL problem is that agents must pre-train on static datasets (Section \ref{section: preliminaries}). This is the realm of offline RL \citep{lange2012batch, levine2020}, where regularisation techniques \citep{kumar2020, kidambi2020, fujimoto2021minimalist} are used to minimise the \textit{distribution shift} between the offline data and online experience \citep{kumar2019stabilizing}. In this work, we only train on high-coverage datasets to isolate the problem of partial observability, so do not require such regularisation, but past work has repurposed these for zero-shot RL \citep{jeen2024zeroshot}.
Standard offline RL methods are trained with respect to one downstream task, so cannot generalise to new tasks at test time, as specified by the zero-shot RL problem. 

\textbf{Goal-conditioned RL} \; For goal-reaching tasks, zero-shot goal generalisation can be achieved with \textit{goal-conditioned} RL (GCRL) \citep{schaul2015, andrychowicz2017hindsight}. Here, policies are trained to reach any goal state from any other state. Past work has focused on constructing useful goal-space encodings, with contrastive \citep{eysenbach2022contrastive}, state-matching \citep{ma2022far}, and hierarchical representations \citep{park2024hiql} proving effective. However, GCRL methods do not reliably generalise to \textit{dense} reward functions that cannot be codified by a goal state,\footnote{Examples include any locomotion task \textit{e.g.} \texttt{Walker}-\texttt{\texttt{run}} in the DeepMind Control Suite.} and so cannot be said to solve the general zero-shot RL problem.


\textbf{Behaviour foundation models} \; To date, BFMs have shown the best zero-shot RL performance because they provide a mechanism for zero-shot generalising to \textit{both} goal-reaching and dense reward functions.\footnote{A formal justification of this statement is left for Section \ref{section: preliminaries}.} They build upon successor representations \citep{dayan1993}, universal value function approximators \citep{schaul2015}, successor features \citep{barreto2017} and successor measures \citep{blier2021learning}. State-of-the-art methods instantiate these ideas as either universal successor features (USFs) \citep{borsa2018, park2024foundation} or forward-backward (FB) representations \citep{touati2021learning, touati2022does, jeen2024zeroshot}. No works have yet explored the zero-shot RL performance of these methods under partial observability.

\vspace{-0.1cm}
\subsection{Partial Observability}
\vspace{-0.1cm}

\textbf{States} \; Most past works assume it is the \textit{state} that is partially observed. This is usually the result of noisy \citep{meng2021memory}, occluded \citep{heess2015memory}, aliased \citep{whitehead1990active}, egocentric \citep{tirumala2024learning} or otherwise unreliable observations. Standard solutions methods use histories of observations and actions to compute \textit{beliefs} over the true state via (approximate) Bayesian inference \citep{cassandra1994acting, kaelbling1998} or via memory-based architectures \citep{schmidhuber1990reinforcement,bakker2001reinforcement, hausknecht2015deep, ha2018}.

\textbf{Dynamics} \; Sometimes, parameters that modulate the underlying \textit{dynamics} change and are not communicated to the agent via the state. Given sets of training and testing dynamics parameters, \textit{generalisation} is a measure of the agent's average-case performance on the test set \citep{packer2018assessing, cobbe2019quantifying}. If the agent trains and tests on the same set of dynamics, \textit{robustness} is a measure of the agent's worst-case performance on this set \citep{nilim2005robust, morimoto2005robust, mankowitz2019robust}. Generalisation can be improved via regularisation \citep{farebrother2018generalization}, data augmentation \citep{tobin2017domain, raileanu2020automatic, ball2021augmented}, or dynamics context modelling \citep{seo2020trajectory, lee2020context}. Robustness can be improved with adversarial dynamics selection \citep{rajeswaran2016epopt, jiang21, rigter2023reward}.

\textbf{Rewards} \; In some cases, the utility of an action for a task may only be partially reflected in the standard one-step reward \citep{minsky1961steps, sutton1984temporal}. Such a situation arises when the reward signal is delayed \citep{arjona2019rudder} or is dependent on the entire trajectory (i.e. episodic) \citep{liu2019off}. These have traditionally been handled with sophisticated techniques that learn surrogate reward functions \citep{raposo2021synthetic, arjona2019rudder}, tune discount factors \citep{fedus2019hyperbolic}, or utilise eligibility traces \citep{xu2020meta}, among other methods.  

Each of the above methods were developed for a specific form of partial observability, but memory-based architectures are, in principle, general enough to solve all of them \citep{kaelbling1998}. Indeed, \cite{ni2021recurrent} find that a standard, but well-implemented, recurrent policy and critic can outperform methods specialised for each setting. Our proposed method (Section \ref{subsection: method}) is heavily informed by this finding, and is designed to be agnostic to the specific way in which partial observability arises.

\section{Preliminaries}\label{section: preliminaries}
\textbf{POMDPs} \; A partially observable Markov decision process (POMDP) $\mathcal{P}$ is defined by $(\mathcal{S}, \mathcal{A}, \mathcal{O}, R, P, O, \mu_0, \gamma)$, where $\mathcal{S}$ is the set of Markov states, $\mathcal{A}$ is the set of actions, $\mathcal{O}$ is the set of observations, and $\mu_0$ is the initial state distribution \citep{aastrom1965, kaelbling1998}. Let $s_t \in \mathcal{S}$ denote the Markov state at time $t$. When action $a_t \in \mathcal{A}$ is executed, the state updates via the transition function $s_{t+1} \sim P(\cdot | s_t, a_t)$, and the agent receives a scalar reward $r_{t+1} \sim R(s_{t+1})$ and observation $o_{t+1} \sim O(\cdot | s_{t+1}, a_t)$. The observation provides only partial information about the underlying Markov state. The agent samples actions from its policy $a_t \sim \pi(\cdot | \tau_{t}^L)$, where $\tau_{t}^L = (a_{t-L}, o_{t-L+1}, \ldots, a_{t-1}, o_t)$ is a \textit{trajectory} of the preceding $L$ observations and actions. We use $\mathcal{T}^L$ to denote the set of all possible trajectories of length $L$.
The policy is optimal in $\mathcal{P}$ if it  maximises the expected discounted future reward \textit{i.e.} $\pi^* = \arg\max_{\pi}\mathbb{E}[\sum_{t\geq 0}\gamma^t R(s_{t+1}) | s_0, a_0, \pi]$, where $\mathbb{E}[\cdot | s_0, a_0, \pi]$ denotes an expectation over state-action sequences $(s_t, a_t)_{t\geq0}$ starting at $(s_0, a_0)$ with $s_t \sim P(\cdot |s_{t-1}, a_{t-1})$ and $a_t \sim \pi(\cdot | \tau^{L}_t)$, and $\gamma \in [0,1]$ is a discount factor.

\textbf{Partially observable zero-shot RL} \; In the standard zero-shot RL problem setting, states are fully observed. For pre-training, the agent has access to a static offline dataset of reward-free transitions $\mathcal{D} = \{(s_i, a_i, s_{i+1})\}_{i=1}^{|\mathcal{D}|}$, generated by an unknown behaviour policy. At test time, a task $R_{\text{test}}$ is revealed by labelling a small number of states in $\mathcal{D}$ to create a new dataset $\mathcal{D}_{\text{labelled}} = \{(s_i, R_{\text{test}}(s_i))\}_{i=1}^{k}$ where typically $k \leq 10,000$. The agent must return a policy for this task with no further planning or learning.

In this work, we consider the extended problem setting of \textit{partially observable zero-shot RL}. Here, the agent has access to an offline pre-training dataset of reward-free length-$L$ trajectories, $\mathcal{D} = \{\tau_{i}^L\}_{i=1}^{|\mathcal{D}|}$, each of which is a sequence of partial observations and actions.
As before, a task $R_{\text{test}}$ is revealed at test time, for which the agent must return a policy. The task is specified by a small dataset of reward-labelled observation-action trajectories, where the reward is assumed to be a function of the final Markov state in the trajectory, $\mathcal{D}_{\text{labelled}} = \{(\tau_{i}^L, R_{\text{test}}(s^L_i))\}_{i=1}^{k}$.

\textbf{Behaviour foundation models} \; We build upon the forward-backward (FB) BFM which predicts successor measures \citep{blier2021learning}. The successor measure $M^{\pi}(s_0, a_0, \cdot)$ over $\mathcal{S}$ is the cumulative discounted time spent in each future state $s_{t+1}$ after starting in state $s_0$, taking action $a_0$, and following policy $\pi$ thereafter. Let $\rho$ be an arbitrary state distribution and $\mathbb{R}^d$ be an embedding space. FB representations are composed of a \textit{forward} model $F: S \times A \times\mathbb{R}^d\to\mathbb{R}^d$, a \textit{backward} model $B: S \to \mathbb{R}^d$, and set of polices $\pi(s, z)_{z \in \mathbb{R}^d}$. They are trained such that:
\begin{align}
    \label{eq:M_approximation}
    M^{\pi_z}(s_0,a_0, X) &\approx \int_X F(s_0,a_0,z)^\top B(s)\rho(\text{d}s) \; \; \qquad \forall \; \; s_0 \in \mathcal{S}, a_0 \in \mathcal{A}, X \subset \mathcal{S}, z \in \mathbb{R}^d, \\
    \label{eq:pi_approximation}
    \pi(s, z) &\approx \arg\max_a F(s,a,z)^\top z \; \qquad \qquad \qquad \qquad \forall \; \; (s,a) \in \mathcal{S} \times \mathcal{A}, z \in \mathbb{R}^d,
\end{align}
where $F(s,a,z)^\top z$ is the $Q$ function (critic) formed by the dot product of forward embeddings with a \textit{task embedding} $z$. During training, candidate task embeddings are sampled from $\mathcal{Z}$, a prior over the embedding space. During evaluation, the test task embeddings are inferred from $\mathcal{D}_{\text{labelled}}$ via:
\begin{equation}\label{equation: z inference}
z_{\text{test}} \approx \mathbb{E}_{(s, R_{\text{test}}(s)) \sim \mathcal{D}_{\text{labelled}}}[R_{\text{test}}(s)B(s)],
\end{equation}
and passed as an argument to the policy.\footnote{Equation \ref{equation: z inference} assumes $\mathcal{D}_{\text{labelled}}$ is drawn from the same distribution as $\mathcal{D}$, that is, it assumes both datasets are produced by the same, exploratory behaviour policy. Deploying a different policy to collect $\mathcal{D}_{\text{labelled}}$ is possible (\textit{e.g.} one learned via Equation \ref{eq:pi_approximation}), but requires minor amendments to Equation \ref{equation: z inference}. We refer the reader to Appendix B.5 of \cite{touati2021learning} for further details.} 

\section{Zero-Shot RL Under Partial Observability}
In this section, we adapt BFMs for the partially observable zero-shot RL setting. In Section \ref{subsection: failure mode of existing methods}, we explore the ways in which standard BFMs fail in this setting. Then, in Section \ref{subsection: method}, we propose new methods that address these failures. We develop our methods in the context of FB, but our proposals are fully compatible with USF-based BFMs. We report their derivation in Appendix \ref{appendix: usf with memory} for brevity.

\subsection{Failure Mode of Existing Methods}\label{subsection: failure mode of existing methods}
\begin{figure}[t]
\centering
    \includegraphics[width=\textwidth]{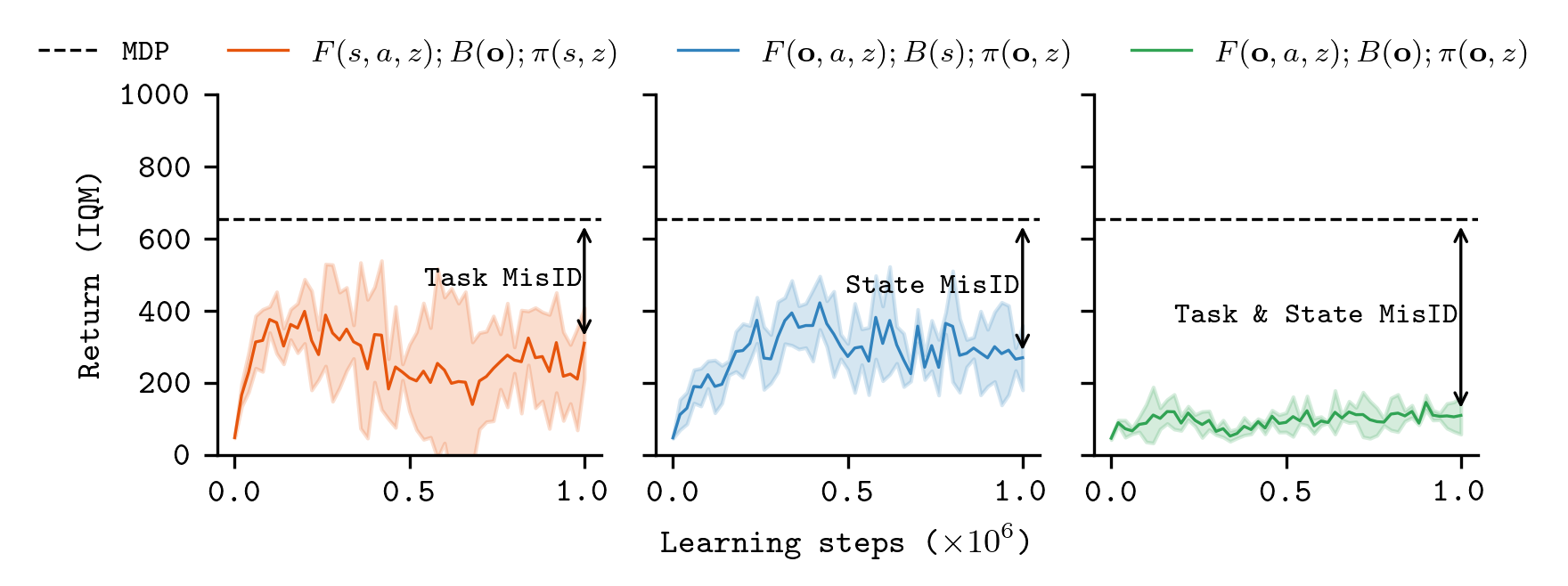}
    \vspace{-0.2cm}
    \caption{\textbf{The failure modes of BFMs under partial observability.} FB's average (IQM) all-task return on \texttt{Walker} when observations are passed to its respective components. Observations are created by adding Gaussian noise to the underlying states. \textit{(Left)} Observations are passed as input to $B$ causing FB to misidentify the task. \textit{(Middle)} Observations are passed as input to $F$ and $\pi$ causing FB to misidentify the state. \textit{(Right)} Observations are passed as input to $F$, $\pi$ and $B$ causing FB to misidentify both the task and state.}
    \label{fig: BFM failure mode}
\end{figure}

FB solves the zero-shot RL problem in two stages. First, a generalist policy is pre-trained to maximise FB's $Q$ functions for all tasks sampled from the prior $\mathcal{Z}$ (Equation \ref{eq:M_approximation}). Second, the test task is inferred from reward-labelled states (Equation \ref{equation: z inference}) and passed to the policy. The first stage relies on an accurate approximation of $F(s,a,z)$ \textit{i.e.} the long-\texttt{run} dynamics of the environment subject to a policy attempting to solve task $z$. The second stage relies on an accurate approximation of $B(s)$ \textit{i.e.} the task associated with reaching state $s$. If the states in $F$ are replaced by observations that only partially characterise the underlying state, then the BFM will struggle to predict the long-\texttt{run} dynamics, $Q$ functions derived from $F$ will be inaccurate, and the policy will not learn optimal sequences of actions. We call this failure mode \textbf{state misidentification} (Figure \ref{fig: BFM failure mode}, middle). Likewise, if the states in $B$ are replaced by partial observations, and the reward function depends on the underlying state (Section \ref{section: preliminaries}), then the BFM cannot reliably find the task $z$ associated with the set of states that maximise the reward function. We call this failure mode \textbf{task misidentification} (Figure \ref{fig: BFM failure mode}, left). The failure modes occur together when both models receive partial observations (Figure \ref{fig: BFM failure mode}, right).

\subsection{Addressing Partial Observability with Memory Models}\label{subsection: method}
In principle, all forms of partial observability can be resolved with \textit{memory models} that compress trajectories into a hidden state that approximates the underlying Markov state (see Section \ref{section: related work} of \citet{ni2021recurrent}). A memory model is a function $f$ that outputs a new hidden state $h_t$ given a past hidden state $h_{t-L-1}$ and trajectory $\tau_{t}^L$:
\begin{equation}\label{equation: generalised memory model}
    h_t = f(\tau_{t}^L, h_{t-L-1}).
\end{equation}
Note that by setting $L=0$, we recover the standard one-step formulation of a recurrent neural network (RNN) \citep{elman1990finding, hochreiter1997, cho2014learning}. RNNs are common choice in past works \citep{wierstra2007policy, zhang2016learning, schmidhuber2019reinforcement}, but more recent works explore structured state space sequence models (S4) \citep{deng2023facing, lu2024structured} and transformers \citep{parisotto2020stabilizing, grigsby2023amago, grigsby2024amago}. In \textit{model-based} partially observable RL, state misidentification is resolved with memory-based dynamics models, and task misidentification is resolved with a memory-based reward models \citep{hafner2019, hafner2019learning, hafner2020mastering, hafner2023mastering}. In \textit{model-free} partially observable RL, the agent does not disentangle the state from the task, so task and state misidentification are resolved together by memory-based critics and policies \citep{ni2021recurrent, meng2021memory}.  
\vspace{-0.1cm}
\subsection{Behaviour Foundation Models with Memory}
We now adapt methods from single-task partially observable RL for BFMs. Standard FB operates on states (Equation \ref{eq:M_approximation}) that are inaccessible under partial observability, so we amend its formulation to operate on trajectories from which the underlying Markov state can be inferred with a memory model. The successor measure $M^\pi(\tau_0^L, \cdot)$ over $\mathcal{T}^L$ is the cumulative discounted time spent in each future trajectory $\tau_{t+1}^L$ starting from trajectory $\tau_0^L$, and following policy $\pi$ thereafter.\footnote{Note that the forward model and backward model can in principle have different context lengths. This is helpful if, for example, we know that the reward, as inferred via the backward model, depends on a shorter history length than would be required to infer the full Markov state via the forward model.} The architectures of the forward model $F$, backward model $B$, and policy $\pi$ are unchanged, but now condition on the hidden states of memory models, rather than on states and actions. They are trained such that
\begin{align}
    \label{eq:M_approximation with memory}
    M^{\pi_z}(\tau_0^L, X) &\approx \int_X F(f_F(\tau_0^L), z)^\top B(f_B(\tau^L))\rho(\text{d}\tau^L) \; \qquad \qquad \forall \; \tau_0^L \in \mathcal{T}^L, X \subset \mathcal{T}^L, z \in \mathbb{R}^d, \notag \\
    \pi(f_{\pi}(\tau^L), z) &\approx \arg\max_a F(f_F(\tau^L),z)^\top z \; \qquad \qquad \qquad \qquad  \; \; \; \;  \forall \;  \tau^L \in \mathcal{T}^L, z \in \mathbb{R}^d.
\end{align}
where $f_F, f_B, f_\pi$ are separate memory models for $F$, $B$, and $\pi$ respectively. Observation and action sequences are zero-padded for all $t - L -1 < 0$; the first hidden state in a sequence is always initialized to zero; and hidden states are dropped as arguments in Equation \ref{eq:M_approximation with memory} for brevity (\textit{c.f.} Equation \ref{equation: generalised memory model}). At test time, task embeddings are found via Equation \ref{equation: z inference}, but with reward-labelled trajectories rather than reward-labelled states:
\begin{equation}
    z_{\text{test}} \approx \mathbb{E}_{(\tau^L, R(s)) \sim \mathcal{D}_{\text{labelled}}}[R_{\text{test}}(s)B(f_B(\tau^L)].
\end{equation}
We refer to the resulting model as \textit{FB with memory} (FB-M).
The full architecture is summarised in Figure \ref{fig: confb architecture}, and implementation details are provided in Appendix \ref{appendix: implementation details}. Also note that our general proposal is BFM-agnostic; we derive the USF-based BFM formulation in Appendix \ref{appendix: usf with memory}. 

\section{Experiments}\label{section: experiments}
\subsection{Setup}\label{subsection: setup}
We evaluate our proposals in two partially observed settings: 1) partially observed states (\textit{i.e.} standard POMDPs), and partially observed changes in dynamics (\textit{i.e.} generalisation \citep{packer2018assessing}). The standard benchmarks for each of these settings only require the agent to solve one task, and so do not allow us to evaluate zero-shot RL capabilities out-of-the-box. As a result, we choose to amend the standard zero-shot RL benchmark, ExORL \citep{yarats2022}, such that it tests zero-shot RL with partially observed states and dynamics changes. 

\textbf{Partially observed states} \; 
We amend two of \cite{meng2021memory}'s partially observed state environments for the zero-shot RL setting: 1) \texttt{noisy} states, where isotropic zero-mean Gaussian noise with variance $\sigma_{\text{noise}}$ is added to the Markov state, and 2) \texttt{flickering} states, where states are dropped (zeroed) with probability $p_{\text{flick.}}$. We set $\sigma_{\text{noise}} = 0.2$ and $p_{\text{flick.}} = 0.2$ following a hyperparameter study (Appendix \ref{appendix: pomdp hyperparameters}). We evaluate on all tasks in the \texttt{Walker}, \texttt{Cheetah} and \texttt{Quadruped} environments.

\textbf{Partially observed changes in dynamics} \;
We amend \cite{packer2018assessing}'s dynamics generalisation tests for the zero-shot RL setting. Environment dynamics are modulated by scaling the mass and damping coefficients in the MuJoCu backend \citep{todorov2012}. The agents are trained on datasets collected from environment instances with coefficients scaled to $\{0.5\times, 1.5\times\}$ their usual values, then evaluated on environment instances with coefficients scaled by $\{1.0\times, 2.0\times\}$. Scaling by $1.0\times$ tests the agent's ability to generalise via \textit{interpolation} within the range seen during training, and scaling by $2.0\times$ tests the agent's ability to generalise via \textit{extrapolation} \citep{packer2018assessing}.


\textbf{Baselines} \; We use two state-of-the-art zero-shot RL methods as baselines: FB \citep{touati2021learning} and HILP \citep{park2024foundation}. We additionally implement a na\"ive baseline called FB-stack whose input is a \textit{stack} of the 4 most recent observations and actions, following \cite{mnih2015}'s canonical protocol. Finally, we use FB trained on the underlying MDP as an oracle policy to give us an upper-bound on expected performance. All methods inherit the default hyperparameters from previous works \citep{touati2022does, park2024foundation, jeen2024zeroshot}.  

\textbf{Datasets} \; We train all methods on datasets collected with an RND behaviour policy \citep{borsa2018} as these are the datasets that elicit best performance on ExORL. The RND datasets used in the partially observed states experiments are taken directly from ExORL. For the partially observed change in dynamics experiments, we collect these datasets ourselves by \texttt{run}ning RND in each of the environments for 5 million learning steps. Our implementation and training protocol exactly match ExORL's.

\textbf{Memory model} \; We use a GRU as our memory model \citep{cho2014learning}. GRUs are the most performant memory model on POPGym \citep{morad2023popgym} which tests partially observed single-task RL methods. We find these results hold for partially observed zero-shot RL too, as discussed in Section \ref{discussion: memory model}.
We set the context length $L = 32$; see Appendix \ref{appendix: context lengths} for a hyperparameter study and further discussion. 

\textbf{Evaluation protocol} \;
We evaluate the cumulative reward achieved by all methods across $5$ seeds, with task scores reported as per the best practice recommendations of \cite{agarwal2021deep}. Concretely, we \texttt{run} each algorithm for 1 million learning steps, evaluating task scores at checkpoints of $20,000$ steps. At each checkpoint, we perform $10$ rollouts, record the score of each, and find the interquartile mean (IQM). We average across seeds at each checkpoint. We extract task scores from the learning step for which the all-task IQM is maximised across seeds. Results are reported with $95\%$ bootstrap confidence intervals in plots and standard deviations in tables. Aggregation across tasks, domains and datasets is always performed by evaluating the IQM. 

\subsection{Partially Observed States}\label{subsection: partially observed states}
\begin{figure}[t]
\centering
    \includegraphics[width=\textwidth]{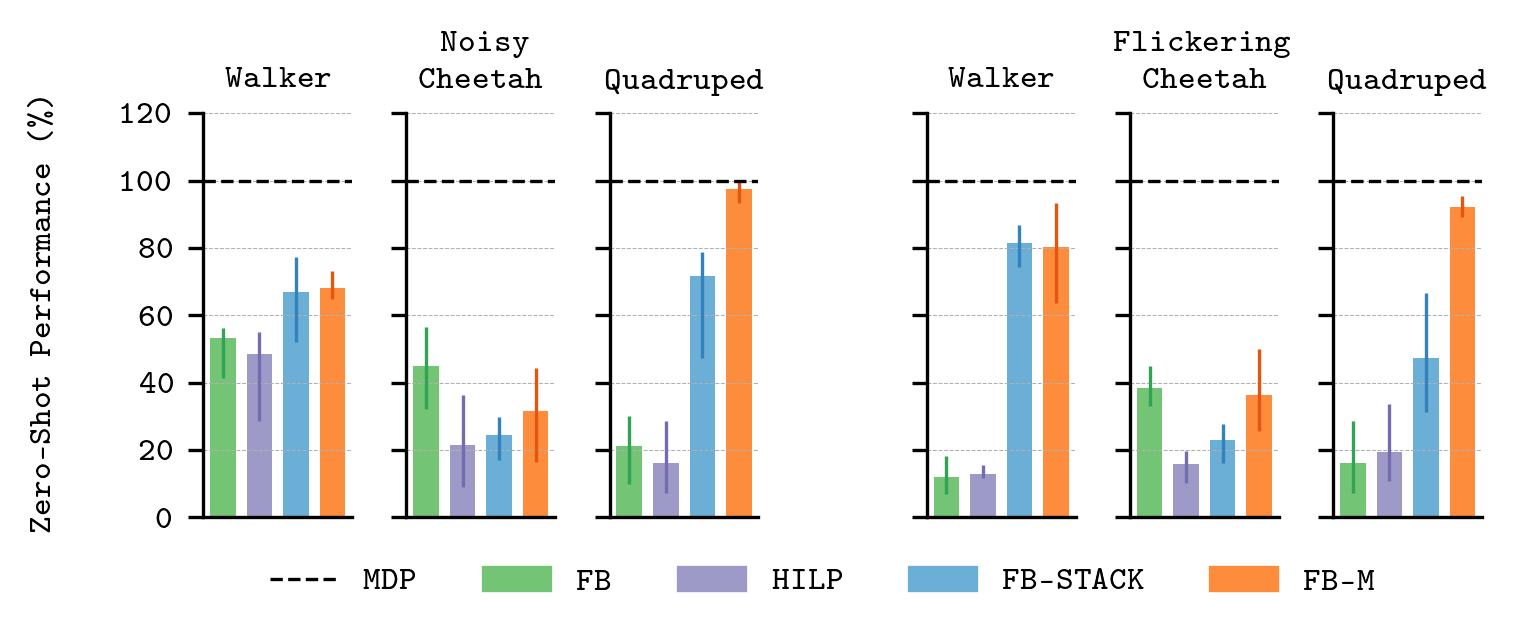}
    \caption{\textbf{Aggregate zero-shot task performance on ExORL with partially observed states.} IQM of task scores across all tasks on \texttt{noisy} and \texttt{flickering} variants of \texttt{Walker}, \texttt{Cheetah} and \texttt{Quadruped}, normalised against the performance of FB in the fully observed environment. 5 random seeds.} 
    \label{fig: results/pomdp-states}
\end{figure}
Figure~\ref{fig: results/pomdp-states} compares the zero-shot performance of all algorithms on our \texttt{noisy} and \texttt{flickering} variants of the standard ExORL environments. Note that these results are aggregated across all tasks in each environment, and 5 random seeds.
The performance of memory-free FB is always far below that of an oracle policy trained on the underlying MDP (dotted line), reaching less than $50\%$ of the oracle value in 5 out of 6 cases, and HILP performs similarly.
Augmenting FB by stacking recent observations mitigates the partial observability problem to some extent on \texttt{Walker} and \texttt{Quadruped}, but it performs worse than memory-free FB on \texttt{Cheetah} . Our proposed approach (FB-M) outperforms this baseline in all settings except \texttt{Walker} where it performs similarly.
The benefit of FB-M is most pronounced for the \texttt{Quadruped} environment where it achieves close to oracle performance. The full results are reported in Table \ref{table: partially observed states} in Appendix \ref{appendix: extended results}.

\subsection{Partially Observed Changes in Dynamics}\label{subsection: partially observed dynamics}
Next, we consider the problem of partially observed dynamics changes in both the interpolation and extrapolation settings.
The results are summarised in Figure~\ref{fig: results/generalisation} and reported in full in Table \ref{table: full changed dynamics} in Appendix \ref{appendix: extended results}.
First, we find that memory-free FB performs well in the interpolation setting, matching the oracle policy in two of the three environments, but less well in the extrapolation setting where it underperforms the oracle in all environments. As in Section \ref{subsection: partially observed states}, HILP is consistently the lowest scoring method in all environments. In general, stacking recent observations (as in FB-stack) harms performance, with scores lower than memory-free FB in 5 out of 6 environment/dynamics settings. FB-M performs similarly to, or better than, all algorithms in all settings. The performance difference is most pronounced in the extrapolation setting on \texttt{Cheetah} and \texttt{Quadruped} where it slightly outperforms the oracle policy. We think this is because the dataset collected under the extrapolation dynamics (with doubled mass and damping coefficients) is less expressive than the standard dynamics because the behaviour policy struggled to cover the state space. Relative differences in the non-MDP results remain valid should this be the case. 
\begin{figure}[t]
\centering
    \includegraphics[width=\textwidth]{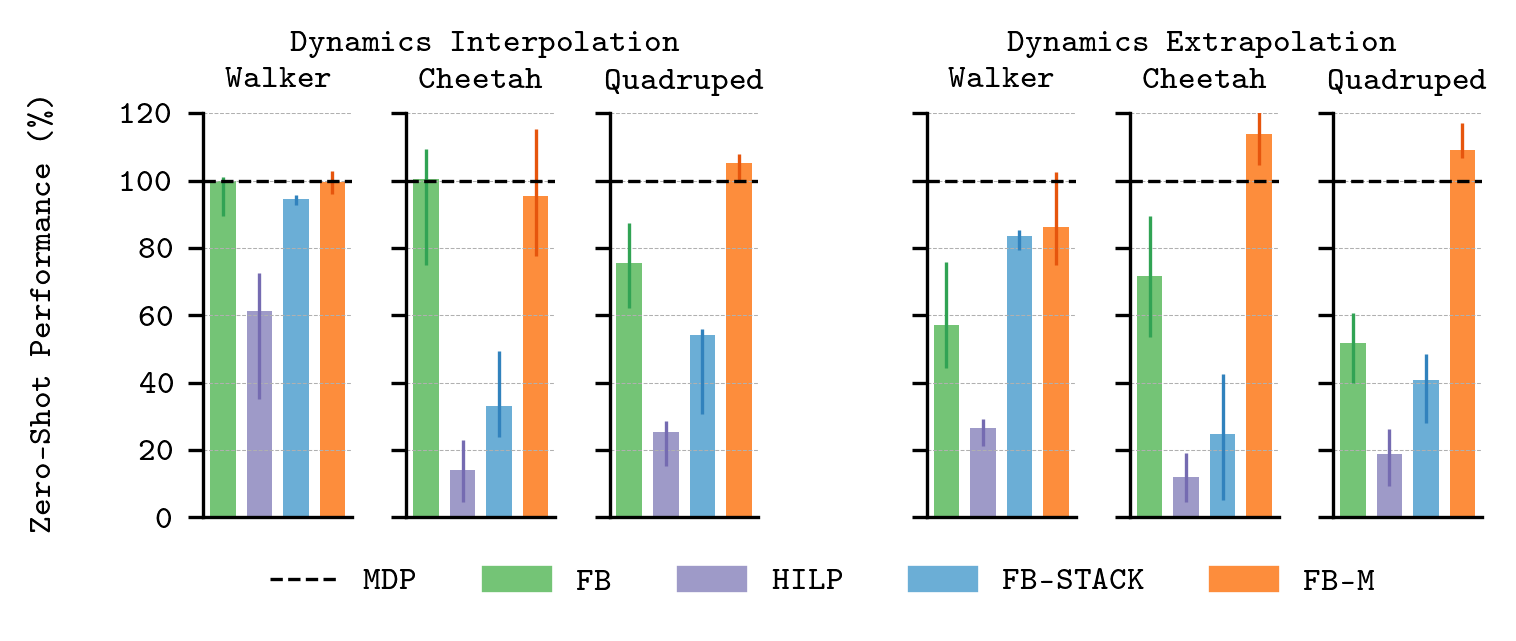}
    \caption{\textbf{Aggregate zero-shot task performance on ExORL with changed dynamics at test time.} IQM of task scores across all tasks when trained on dynamics where mass and damping coefficients are scaled to \{$0.5\times$, $1.5\times$\} their usual values and evaled on \{$1.0\times$, $2.0\times$\} their usual values, normalised against the performance of FB in the fully observed environment. To solve the test dynamics with $1.0\times$ scaling the agent must interpolate within the training set; to solve the test dynamics with $2.0\times$ scaling the agent must extrapolate from the training set.}
    \label{fig: results/generalisation}
\end{figure}
\vspace{-0.1cm}
\section{Discussion and Limitations}\label{section: discussion}
\subsection{Memory Model Choice} \label{discussion: memory model}
Our method uses GRUs as memory models, but much recent work has shown that transformers \citep{vaswani2017attention} and structured state-space models \citep{gu2021efficiently} outperform GRUs in natural language processing \citep{brown2020language}, computer vision \citep{dosovitskiy2020image}, and model-based RL \citep{deng2023facing}. In this section, we explore whether these findings hold for the zero-shot RL setting. We compare FB-M with GRU memory models to FB-M with transformer and diagonalised S4 (S4d) memory models \citep{gu2022parameterization}. We follow \cite{morad2023popgym} in restricting each model to a fixed hidden state size, rather than a fixed parameter count, to ensure a fair comparision. Concretely, we allow each model a hidden state size of $32^2 = 1024$ dimensions. Full implementation details are provided in Appendix \ref{appendix: implementation details}. We evaluate each method in the three variants of \texttt{Walker} \texttt{flickering} used in Section \ref{subsection: failure mode of existing methods} \textit{i.e.} where only the inputs to $F$ and $\pi_z$ are observations, only inputs to $B$ are observations, and where inputs to all models are observations.

\begin{figure}[ht]
\centering
    \includegraphics[width=0.9\textwidth]{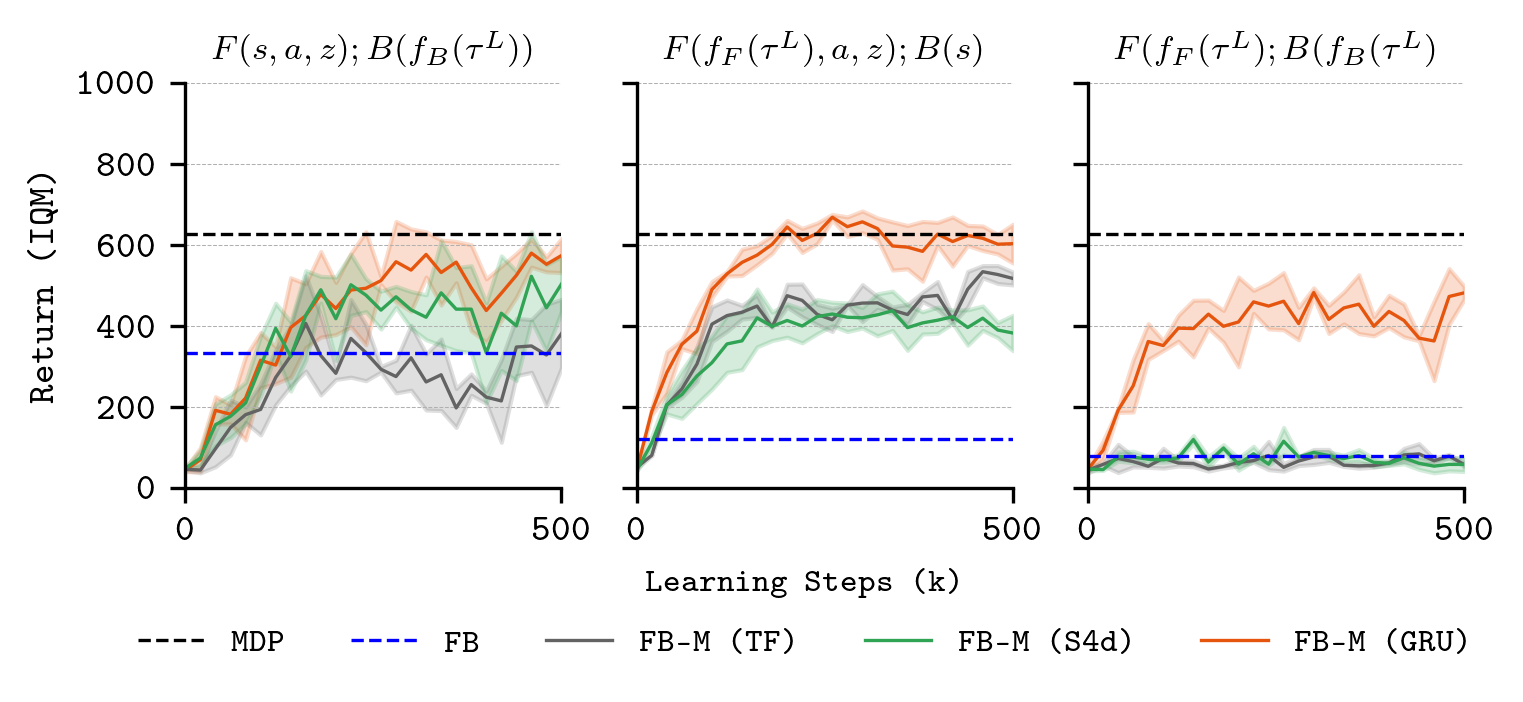}
    \vspace{-0.3cm}
    \caption{\textbf{Aggregate zero-shot task performance of FB-M with different memory models.} IQM of task scores across all tasks on \texttt{Walker flickering}. \textit{(Left)} Observations are passed only to a memory-based backward model; the forward model and policy are memory-free. \textit{(Middle)} Observations are passed only to the forward model and policy; the backward model is memory-free. \textit{(Right)} Observations are passed to all models.}
    \label{fig: results/memory-models}
\end{figure}

Our results are reported in Figure \ref{fig: results/memory-models}. We find that the performance of FB-M is reduced in all cases when a transformer or S4 memory model is used instead of a GRU. This corroborates \cite{morad2023popgym}'s findings that the GRU is the most performant memory model for single-task partially observed RL. Perhaps most crucially, we find that training collapses when \textit{both} $F$ and $B$ are non-GRU memory models, despite non-GRU memory models performing reasonably when added to \textit{only} $F$ or $B$, suggesting that the combined representation $M(\tau^L, \tau^L_+) \approx F(f_F(\tau^L))^\top B(f_B(\tau_+^L))$ is degenerate. Better understanding this failure mode is important future work.

\subsection{Datasets}
As outlined in Section \ref{subsection: setup}, we train all methods on datasets pre-collected with RND \citep{borsa2018} which is a highly exploratory algorithm designed for maximising data heterogeneity. However, deploying such an algorithm in any real setting may be costly, time-consuming or dangerous. As a result, our proposals are more likely to be trained on real-world datasets that are smaller and more homogeneous. It is not clear how our specific proposals will interact with such datasets. If, for example, the dataset only represents parts of the state space from which the dynamics cannot be well-inferred, because it was collected from a robot with limited freedom of movement, then we would expect our proposals to struggle. Indeed, with poor coverage of the state-action space, we would expect to see the OOD pathologies seen in the single-task offline RL setting \citep{kumar19, levine2020}. That said, the proposals of \cite{jeen2024zeroshot} for conducting zero-shot RL from less diverse datasets could be integrated into our proposals easily, and may help.

\section{Conclusion}

In this paper, we explored how the performance of BFMs degrades when subjected to certain types of partial observability. We introduce memory-based BFMs that condition $F$, $B$ and $\pi_z$ on trajectories of observation-action pairs, and show they go some way to remedying state and task misidentification. We evaluated our proposals on a suite of partially observed zero-shot RL problems, where the observations passed to the agent are noisy, dropped randomly or do not communicate a change in the underlying dynamics, and showed improved performance over memory-free baselines. We found the GRU to be the most performant memory model, and showed that transformers and s4 memory models cannot be trained stably at our scale. We believe our proposals represent a further step towards the real-world deployment of zero-shot RL methods.



\newpage
\bibliography{bibliography}
\bibliographystyle{rlj}

\appendix
\newpage
\beginSupplementaryMaterials

\startcontents[appendices]
\printcontents[appendices]{l}{0}{\setcounter{tocdepth}{2}}
\newpage

\section{Experimental Details}\label{appendix: experimental details}
\subsection{ExORL}
We consider 3 environments (three locomotion and one goal-directed) from the ExORL benchmark \citep{yarats2022} which is built atop the DeepMind Control Suite \citep{tassa2018}. Environments are visualised here: \url{https://www.youtube.com/watch?v=rAai4QzcYbs}. The domains are summarised in Table \ref{table: domain summary}.

\textbf{\texttt{Walker}} \; A two-legged robot required to perform locomotion starting from bent-kneed position. The observation and action spaces are 24 and 6-dimensional respectively, consisting of joint torques and positions. ExORL provides 4 tasks \texttt{stand, \texttt{walk}, \texttt{run}} and \texttt{flip.} The reward function for \texttt{stand} motivates straightened legs and an upright torso; \texttt{\texttt{walk}} and \texttt{\texttt{run}} are  supersets of \texttt{stand} including reward for small and large degrees of forward velocity; and \texttt{flip} motivates angular velocity of the torso after standing. Rewards are dense.

\textbf{\texttt{Quadruped}} \; A four-legged robot required to perform locomotion inside a 3D maze. The observation and action spaces are 84 and 12-dimensional respectively, consisting of joint torques and positions. We evaluate on 4 tasks \texttt{stand, \texttt{run}, \texttt{walk}} and \texttt{jump.} The reward function for \texttt{stand} motivates a minimum torso height and straightened legs; \texttt{\texttt{walk}} and \texttt{\texttt{run}} are  supersets of \texttt{stand} including reward for small and large degrees of forward velocity; and \texttt{jump} adds a term motivating vertical displacement to stand. Rewards are dense.

\begin{table}[b]
\caption{\textbf{ExORL domain summary.} \textit{Dimensionality} refers to the relative size of state and action spaces. \textit{Type} is the task categorisation, either locomotion (satisfy a prescribed behaviour until the episode ends) or goal-reaching (achieve a specific task to terminate the episode). \textit{Reward} is the frequency with which non-zero rewards are provided, where dense refers to non-zero rewards at every timestep and sparse refers to non-zero rewards only at positions close to the goal. {\color[HTML]{009901} Green} and {\color[HTML]{CB0000} red} colours reflect the relative difficulty of these settings.}

\label{table: domain summary}
\centering
\begin{tabular}{llll}
\toprule
\textbf{Environment}                        & \textbf{Dimensionality}     & \textbf{Type}                        & \textbf{Reward}               \\
 \midrule
{\color[HTML]{000000} \texttt{Walker}}& {\color[HTML]{009901} Low}  & {\color[HTML]{000000} Locomotion}    & {\color[HTML]{009901} Dense}  \\
\texttt{Quadruped}                              & {\color[HTML]{CB0000} High} & Locomotion                           & {\color[HTML]{009901} Dense}                         \\
 
\texttt{Cheetah}                                   & {\color[HTML]{009901} Low} & Locomotion                        & {\color[HTML]{009901} Dense}                       \\
\bottomrule
\end{tabular}
\end{table}

\textbf{\texttt{Cheetah}} \; A \texttt{run}ning two-legged robot.  The observation and action spaces are 17 and 6-dimensional respectively, consisting of positions of robot joints. We evaluate on 4  tasks: \texttt{\texttt{walk}, \texttt{walk backward}, \texttt{run}} and \texttt{\texttt{run backward}.} Rewards are linearly proportional either a forward or backward velocity--2 m/s for \texttt{walk} and 10 m/s for \texttt{run}.

\subsection{POMDP Hyperparameters}\label{appendix: pomdp hyperparameters}
\begin{figure}[h]
\centering
    \includegraphics[width=0.8\textwidth]{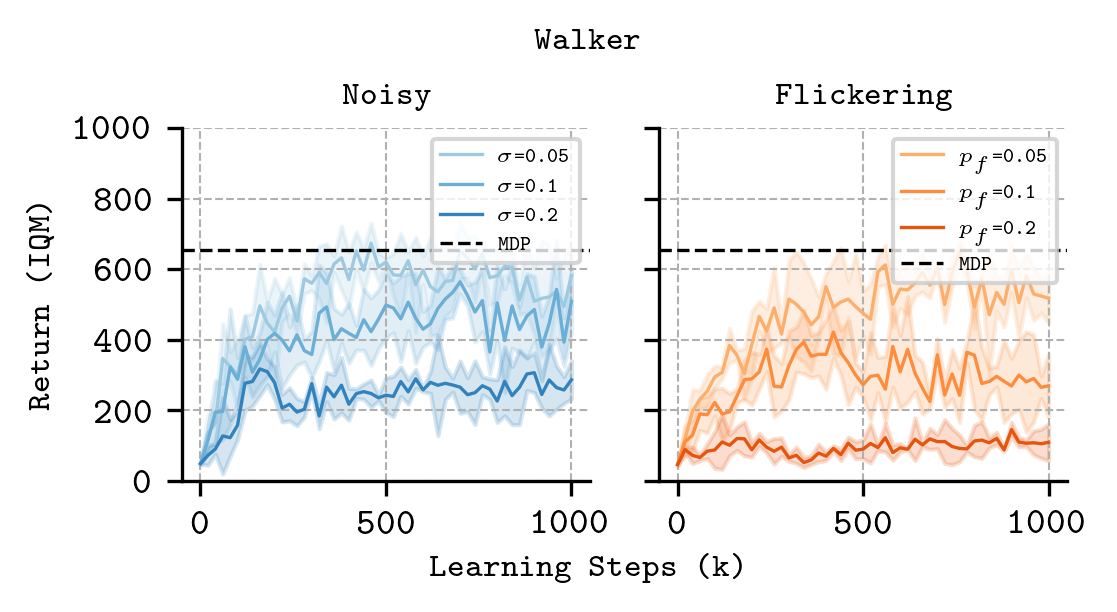}
    \caption{\textbf{POMDP hyperparameter sweep.} We evaluate the performance of standard FB on \texttt{Walker} when the states are noised according to $\sigma \in \{0.05, 0.1, 0.2\}$ and dropped according to $p_f \in \{0.05, 0.1, 0.2\}$.}
    \label{fig: appendix/pomdp hyperparameters}
\end{figure}

The \texttt{noisy} and \texttt{flickering} amendments to standard ExORL environments (Section \ref{section: experiments}) have associated hyperparameters $\sigma$ and $p_f$. Hyperparameter $\sigma$ is the variance of the 0-mean Gaussian from which noise is sampled before being added to the state, and $p_f$ is the probability that state $s$ is dropped (zeroed) at time $t$. In Figure \ref{fig: appendix/pomdp hyperparameters} we sweep across three valued of each in $\{0.05, 0.1, 0.2\}$. From these findings we set $\sigma=0.2$ and $p_f = 0.2$ in the main experiments 

\subsection{Computational Resources}\label{appendix: computational resources}
We train our models on NVIDIA A100 GPUs. One run of FB, FB-stack and HILP on one domain (for all tasks) takes approximately 6 hours on one GPU. One \texttt{run} of the FB-M on one domain (for all tasks) on one GPU in approximately 20 hours. As a result, our core experiments on the ExORL benchmark used approximately 65 GPU days of compute. 

\section{Universal Successor Features with Memory}\label{appendix: usf with memory}
USFs require access to a feature map $\varphi: S \mapsto \mathbb{R}^d$ that maps states into an embedding space in which the reward is assumed to be linear \textit{i.e.} $R(s) = \varphi(s)^\top z$ with \textit{weights} $z \in \mathbb{R}^d$ representing a task \citep{barreto2017, borsa2018}. The USFs $\psi: S \times A \times \mathbb{R}^d \rightarrow \mathbb{R}^d$ are defined as the discounted sum of future features subject to a task-conditioned policy $\pi_z$, and are trained such that
\begin{align}
    \psi(s_0, a_0, z) &= \mathbb{E}\left[\sum_{t \geq 0} \gamma^t \varphi(s_{t+1}) | s_0, a_0, \pi_z \right] \qquad\; \; \; \; \; \qquad \forall \; s_0 \in S, a_0 \in A, z \in \mathbb{R}^d \\
    \pi(s, z) &\approx \arg\max_a \psi(s,a,z)^\top z, \; \; \qquad \qquad \qquad \qquad \forall \; s \in S, a \in A,  z \in \mathbb{R}^d.
\end{align}
During training candidate task weights are sampled from $\mathcal{Z}$; during evaluation, the test task weights are found by regressing labelled states onto the features: 
\begin{equation}\label{equation: usf task inference}
z_{\text{test}} \approx \argminA_z \mathbb{E}_{s \sim \mathcal{D}_{\text{test}}}[(R_{\text{test}}(s) - \varphi(s)^\top z)^2],
\end{equation}
before being passed to the policy. The features can be learned with Hilbert representations \citep{park2024foundation}, laplacian eigenfunctions, or contrastive methods \citep{touati2022does}.

We define \textit{memory-based} USFs as the discounted sum of future features extracted from the memory model's hidden state, subject to a memory-based policy $\pi_z(f_\pi(\tau^L)$:
\begin{align}
\psi(\tau_0^L, z) &= \mathbb{E}\left[\sum_{t \geq 0} \gamma^t \varphi(f_\psi(\tau_{t+1}^L)) \; | \;  \tau_0^L, \pi_z \right] \; \; \; \qquad \qquad  \forall \; \tau_0^L \in \mathcal{T}, z \in \mathbb{R}^d \\
\pi(f_\pi(\tau^L), z) &\approx \arg\max_a \psi(f_\psi(\tau^L),z)^\top z \; \; \; \qquad \qquad \qquad \; \forall \; \tau^L \in \mathcal{T},  z \in \mathbb{R}^d, 
\end{align}
where $f_\psi$ and $f_\pi$ are seperate memory models for $\psi$ and $\pi$, and the previous hidden state $h_{t-L-1}$ is dropped as an argument for brevity (\textit{c.f.} Equation \ref{equation: generalised memory model}). At test time, task embeddings are found via Equation \ref{equation: usf task inference}, but this time with reward-labelled trajectories rather than reward-labelled states:
\begin{equation}\label{equation: usf-mem task inference}
z_{\text{test}} \approx \argminA_z \mathbb{E}_{(\tau^L, R(s)) \sim \mathcal{D}_{\text{labelled}}}[(R_{\text{test}}(s) - \varphi(f_\psi(\tau^L)^\top z)^2],
\end{equation}
before being passed to the policy.

\section{Implementation Details}\label{appendix: implementation details}

\subsection{FB-M}
\textbf{Memory Models $f_F(\tau^L)$, $f_B(\tau^L)$ and $f_{\pi}(\tau^L)$} \; FB-M has separate memory models for the forward model $f_F$, backward model $f_B$ and policy $f_{\pi}$ following the findings of \citep{ni2021recurrent}, but their implementations are identical. Trajectories of observation-action pairs are preprocessed by one-layer feedforward MLPs that embed their inputs into a 512-dimensional space. The memory model is a GRU whose hidden state is initialised as zeros and updated sequentially by processing each embedding in the trajectory. For the experiments in Section \ref{discussion: memory model} we additionally use transformer \cite{vaswani2017attention} and s4 memory models \cite{gu2021efficiently}. Our transformer uses \textit{FlashAttention} \citep{dao2022flashattention} for faster inference, and we use diagonalised s4 (s4d) \citep{gu2022parameterization} rather than standard s4 because of its improved empirical performance on sequence modelling tasks.

\textbf{Forward Model $F(f_F(\tau^L), z)$} \; The forward model takes the final hidden state from $f_F$ and concatenates it with a preprocessed embedding of the most recent observation-task pair $(o, z)$, following the standard FB convention \cite{touati2021learning}. This vector is passed through a final feedforward MLP $F$ which outputs a $d$-dimensional embedding vector. 

\textbf{Backward Model $B(f_B(\tau^L))$} \; The backward model takes the final hidden state from $f_B$ passed it through a two-layer feedforward MLP that outputs a $d$-dimensional embedding vector.

\textbf{Actor $\pi(f_{\pi}(\tau^L), z)$} \; The actor takes the final hidden state from $f_\pi$ and concatenates it with a preprocessed embedding of the most recent observation-task pair $(o, z)$, following the standard FB convention \cite{touati2021learning} This vector is passed through a final feedforward MLP which outputs an $a$-dimensional vector, where $a$ is the action-space dimensionality. A \texttt{Tanh} activation is used on the last layer to normalise their scale. As per \citep{fujimoto2019}'s recommendations, the policy is smoothed by adding Gaussian noise $\sigma$ to the actions during training.

\begin{figure}[t]
\centering
    \includegraphics[width=\textwidth]{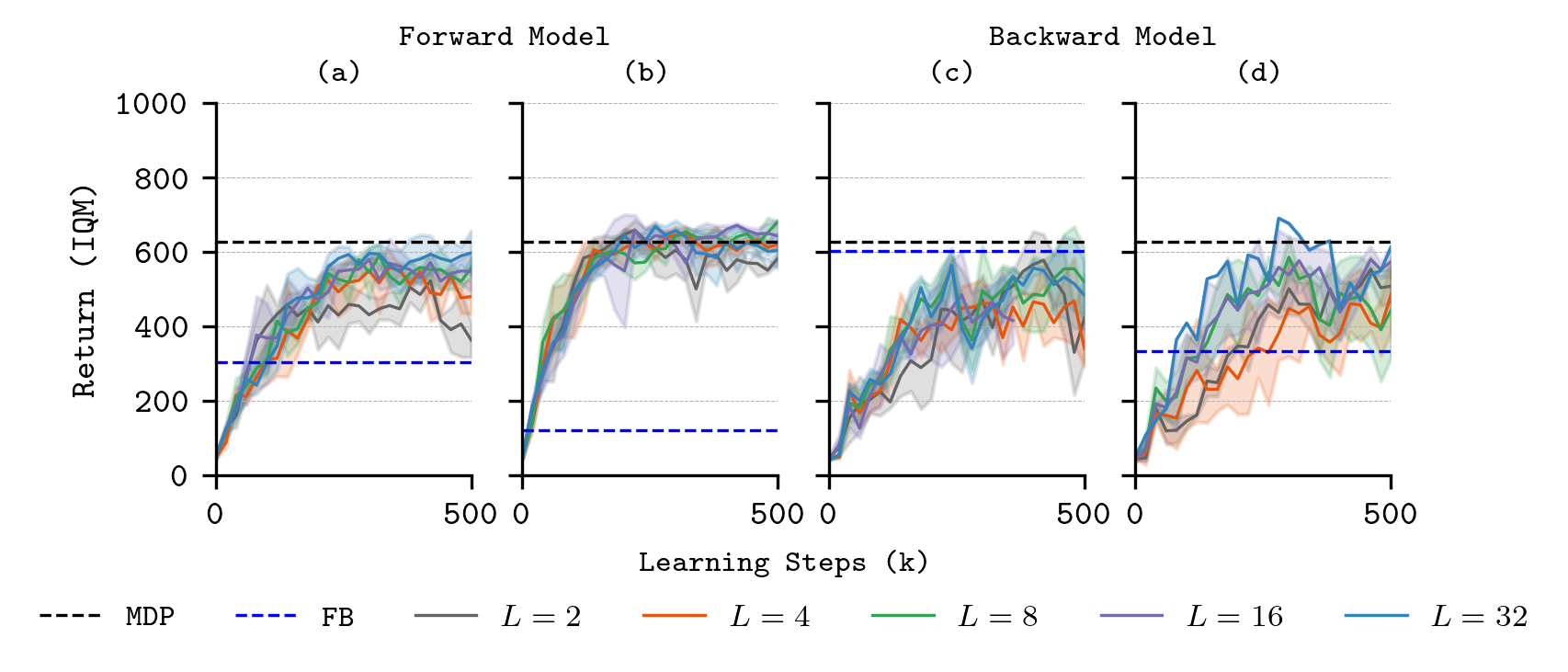}
    \caption{\textbf{Hyperparameter sweep over context length $L$.} We evaluate the performance of FB-M with GRU memory model on \texttt{Walker} \texttt{noisy} ((a) and (c) and \texttt{Walker} \texttt{flickering} ((b) and (d)). When we sweep over the forward model's context length, we pass states to the backward model and keep it memory-free; when we sweep over the backward model's context length we pass states to the forward model and policy and keep them memory-free.}
    \label{fig: appendix/context sweep}
\end{figure}

\subsection{Context Lengths}\label{appendix: context lengths}
The context length $L$ of both the $F/\pi_z$ and $B$ is an important hyperparameter. When adding memory to actors or critics, it is standard practice to parallelise  training across batched trajectories of fixed $L$ (zero-padded for all $t < L$), yet condition the policy on the entire episode history during evaluation with recurrent hidden states. If $L$ is chosen to be less than the maximum episode length, as is often required with limited compute, a shift between the training and evaluation distributions is inevitable. Though this does not tend to harm performance significantly \citep{hausknecht2015deep}, the aim is generally to maximise $L$ subject to available compute.
The Markov states of different POMDPs will require different $L$, but longer $L$ increases training time and risks decreased training stability. In Figure \ref{fig: appendix/context sweep} we sweep across $L \in \{2, 4, 8, 16, 32\}$ for both $F/\pi_z$ and $B$. In general, we see small increases in performance for increased context length, and choose $L = 32$ for our main experiments.  
\subsection{FB and HILP}
\begin{figure}[t]
\centering
    \includegraphics[width=\textwidth]{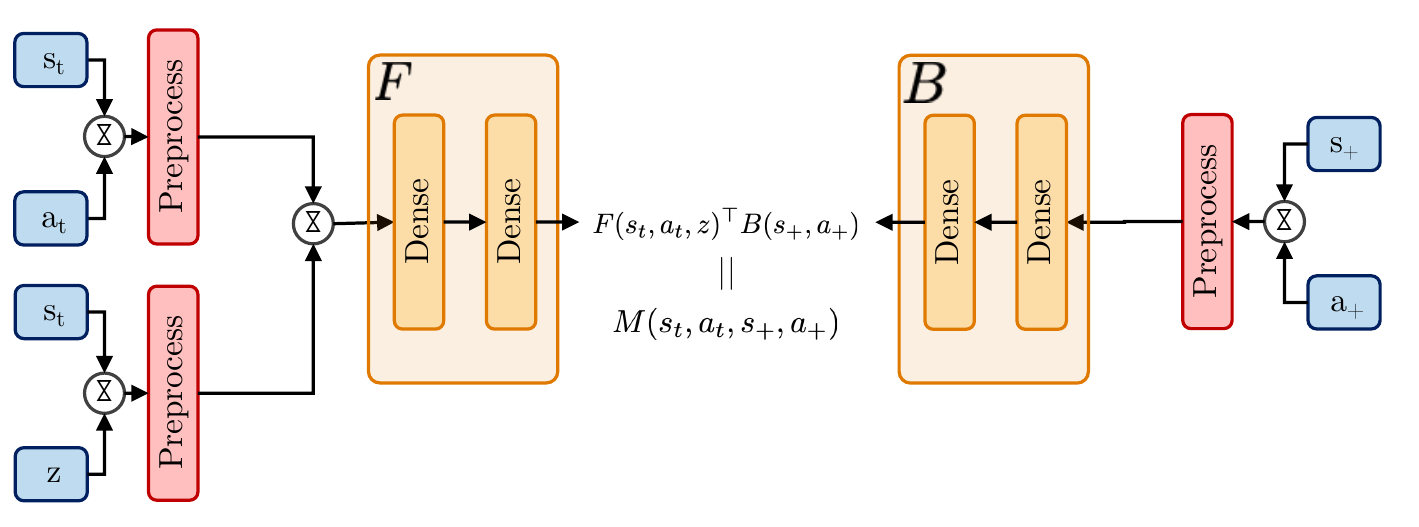}
    \caption{\textbf{BFMs \textit{without} memory}. FB is optimised in a standard actor critic setup \citep{konda1999actor}. The policy $\pi$ selects an action {\color{action-purple}{$a_t$}} conditioned on the current observation $o_t$, and the task vector $z$. The $Q$ function formed by the USF $\psi$ evaluates the action $a_t$ given the current observation $o_t$ and task $z$.}
    \label{fig: fb architecture}
\end{figure}
FB and HILP follow the implementations by \citep{park2024foundation} which follow \citep{touati2022does}, other than the batch size which we reduce from 1024 to 512 to reduce the computational expense of each \texttt{run} without limiting performance as per \citep{jeen2024zeroshot}. Hyperparameters are reported in Table \ref{table: fb hyperparameters}. An illustration of a standard FP architecture is provided in Figure \ref{fig: fb architecture}, for comparison with the FP with memory architecture in Figure \ref{fig: confb architecture}.

\textbf{Forward Model $F(o_t, a_t, z)$ / USF $\psi(o_t, a_t, z)$} \; Observation-action pairs $(o, a)$ and observation-task pairs $(o, z)$ are preprocessed and concatenated before being passed through a final feedforward MLP $F$ / $\psi$ which outputs a $d$-dimensional embedding vector.

\textbf{Backward Model $B(o_t)$ / Feature Embedder $\varphi(o_t)$} \; Observations are preprocessed then passed to the backward model $B$ / feature embedder $\varphi$ which is a two-layer feedforward MLP that outputs a $d$-dimensional embedding vector. 

\textbf{Actor $\pi(o_t, z)$} \; Observations $(o_t)$ and observation-task pairs $(o_t, z)$ are preprocessed by one-layer and concatenated before being passed through a final feedforward MLP which outputs a $a$-dimensional vector, where $a$ is the action-space dimensionality. A \texttt{Tanh} activation is used on the last layer to normalise their scale. As per \citep{fujimoto2019}'s recommendations, the policy is smoothed by adding Gaussian noise $\sigma$ to the actions during training.

\textbf{Misc.} \; Layer normalisation \citep{ba2016layer} and \texttt{Tanh} activations are used in the first layer of all MLPs to standardise the inputs.

\begin{table}\caption{\textbf{Hyperparameters for all BFMs.}}
\label{table: fb hyperparameters}
\centering
\scalebox{1}{
\begin{tabular}{lc}
\toprule
Hyperparameter                       & Value       \\ \midrule
Latent dimension $d$                          & 50     \\
$F$ / $\psi$ dimensions                          & (1024, 1024)                 \\
$B$ dimensions                            & (512, 512)                    \\
Preprocessor dimensions &  (512, 512) \\
Transformer heads & 4 \\ 
Transformer / S4d model dimension & 32 \\ 
GRU dimensions & (512, 512) \\ 
Context length $L$ & 32  \\ 
Frame stacking (FB \& HILP) & 4 \\
Std. deviation for policy smoothing $\sigma$  & 0.2                  \\
T\texttt{run}cation level for policy smoothing         & 0.3                  \\
Learning steps                                & 1,000,000 \\
Batch size                                    & 512                  \\
Optimiser                                     & Adam \citep{kingma2014adam}  \\
Learning rate                                 & 0.0001               \\ 
Discount $\gamma$                             & 0.98 \\
Activations (unless otherwise stated)         & ReLU                 \\
Target network Polyak smoothing coefficient & 0.01                 \\
$z$-inference labels                          & 10,000              \\
$z$ mixing ratio                              & 0.5                   \\
HILP representation discount factor & 0.98 \\
HILP representation expectile & 0.5 \\
HILP representation target smoothing coefficient & 0.005 \\ \bottomrule
\end{tabular}
}
\end{table}

\subsubsection{\texorpdfstring{$z$}{z} Sampling}\label{appendix: z sampling}
BFMs require a method for sampling the task vector $z$ at each learning step. \citep{touati2022does} employ a mix of two methods, which we replicate:
\begin{enumerate}
    \item Uniform sampling of $z$ on the hypersphere surface of radius $\sqrt{d}$ around the origin of $\mathbb{R}^d$,
    \item Biased sampling of $z$ by passing states $s \sim \mathcal{D}$ through the backward model $z = B(s)$. This also yields vectors on the hypersphere surface due to the $L2$ normalisation described above, but the distribution is non-uniform.
\end{enumerate}
We sample $z$ 50:50 from these methods at each learning step.

\subsection{Code References}
This work was enabled by: Python \citep{sanner1999python}, NumPy \citep{numpy}, PyTorch \citep{torch}, Pandas \citep{pandas} and Matplotlib \citep{matplotlib}.


\section{Extended Results}\label{appendix: extended results}
We report a full breakdown of our results summarised in Sections \ref{subsection: partially observed states} and \ref{subsection: partially observed dynamics}. Table \ref{table: partially observed states} reports results on our partially observed states experiments and Table \ref{table: full changed dynamics} reports results on our changed dynamics experiments.

\begin{table}
\centering
\caption{\textbf{Full results on partially observed states (5 seeds).} For each dataset-domain pair, we report the score at the step for which the all-task IQM is maximised when averaging across 5 seeds $\pm$ the standard deviation.}
\label{table: partially observed states}
\scalebox{0.80}{
\begin{tabular}{lllllll|l}
\toprule
 \texttt{Environment} & \texttt{Occlusion} & \texttt{Task}  & \texttt{FB} & \texttt{HILP} & \texttt{FB-stack} & \texttt{FB-M (ours)} & \texttt{MDP} \\
\midrule \addlinespace
\multirow[c]{10}{*}{\texttt{Cheetah}} & \multirow[c]{5}{*}{\texttt{flickering}} &\texttt{overall} & $182 \pm \text{\tiny{$25$}}$ & $75 \pm \text{\tiny{$19$}}$ & $109 \pm \text{\tiny{$26$}}$ & $173 \pm \text{\tiny{$51$}}$ & $474 \pm \text{\tiny{$129$}}$ \\
 &  & \texttt{run} & $54 \pm \text{\tiny{$14$}}$ & $31 \pm \text{\tiny{$13$}}$ & $25 \pm \text{\tiny{$16$}}$ & $75 \pm \text{\tiny{$37$}}$ & $183 \pm \text{\tiny{$59$}}$ \\
 &  & \texttt{run backward} & $52 \pm \text{\tiny{$23$}}$ & $8 \pm \text{\tiny{$5$}}$ & $31 \pm \text{\tiny{$11$}}$ & $55 \pm \text{\tiny{$15$}}$ & $176 \pm \text{\tiny{$53$}}$ \\
 &  & \texttt{walk} & $279 \pm \text{\tiny{$65$}}$ & $184 \pm \text{\tiny{$66$}}$ & $186 \pm \text{\tiny{$103$}}$ & $306 \pm \text{\tiny{$175$}}$ & $726 \pm \text{\tiny{$194$}}$ \\
 &  & \texttt{walk backward} & $293 \pm \text{\tiny{$87$}}$ & $61 \pm \text{\tiny{$21$}}$ & $178 \pm \text{\tiny{$67$}}$ & $278 \pm \text{\tiny{$70$}}$ & $812 \pm \text{\tiny{$217$}}$ \\
\addlinespace\cline{2-8}\addlinespace
 & \multirow[c]{5}{*}{\texttt{noisy}} & \texttt{overall} & $213 \pm \text{\tiny{$53$}}$ & $102 \pm \text{\tiny{$59$}}$ & $116 \pm \text{\tiny{$26$}}$ & $150 \pm \text{\tiny{$59$}}$ & $474 \pm \text{\tiny{$129$}}$ \\
 &  & \texttt{run} & $65 \pm \text{\tiny{$9$}}$ & $28 \pm \text{\tiny{$30$}}$ & $34 \pm \text{\tiny{$11$}}$ & $34 \pm \text{\tiny{$17$}}$ & $183 \pm \text{\tiny{$59$}}$ \\
 &  & \texttt{run backward} & $68 \pm \text{\tiny{$35$}}$ & $24 \pm \text{\tiny{$23$}}$ & $41 \pm \text{\tiny{$14$}}$ & $57 \pm \text{\tiny{$28$}}$ & $176 \pm \text{\tiny{$53$}}$ \\
 &  & \texttt{walk} & $340 \pm \text{\tiny{$62$}}$ & $228 \pm \text{\tiny{$98$}}$ & $186 \pm \text{\tiny{$45$}}$ & $199 \pm \text{\tiny{$88$}}$ & $726 \pm \text{\tiny{$194$}}$ \\
 &  & \texttt{walk backward} & $415 \pm \text{\tiny{$170$}}$ & $133 \pm \text{\tiny{$99$}}$ & $214 \pm \text{\tiny{$70$}}$ & $283 \pm \text{\tiny{$142$}}$ & $812 \pm \text{\tiny{$217$}}$ \\
\addlinespace\cline{1-8}\addlinespace
 \multirow[c]{10}{*}{\texttt{Quadruped}} & \multirow[c]{5}{*}{\texttt{flickering}} & \texttt{overall} & $117 \pm \text{\tiny{$68$}}$ & $140 \pm \text{\tiny{$75$}}$ & $345 \pm \text{\tiny{$120$}}$ & $673 \pm \text{\tiny{$19$}}$  & $729 \pm \text{\tiny{$6$}}$ \\
 &  & \texttt{jump} & $174 \pm \text{\tiny{$93$}}$ & $163 \pm \text{\tiny{$142$}}$ & $377 \pm \text{\tiny{$112$}}$ & $771 \pm \text{\tiny{$29$}}$ & $737 \pm \text{\tiny{$21$}}$ \\
 &  & \texttt{run} & $63 \pm \text{\tiny{$151$}}$ & $95 \pm \text{\tiny{$70$}}$ & $284 \pm \text{\tiny{$114$}}$ & $478 \pm \text{\tiny{$14$}}$ & $504 \pm \text{\tiny{$13$}}$ \\
 &  & Stand & $65 \pm \text{\tiny{$130$}}$ & $98 \pm \text{\tiny{$115$}}$ & $460 \pm \text{\tiny{$188$}}$ & $950 \pm \text{\tiny{$14$}}$ & $955 \pm \text{\tiny{$36$}}$ \\
 &  & \texttt{walk} & $34 \pm \text{\tiny{$65$}}$ & $181 \pm \text{\tiny{$81$}}$ & $291 \pm \text{\tiny{$95$}}$ & $487 \pm \text{\tiny{$51$}}$ & $749 \pm \text{\tiny{$57$}}$ \\
\addlinespace\cline{2-8}\addlinespace
 & \multirow[c]{5}{*}{\texttt{noisy}} & \texttt{overall} & $155 \pm \text{\tiny{$63$}}$ & $117 \pm \text{\tiny{$68$}}$ & $522 \pm \text{\tiny{$111$}}$ & $711 \pm \text{\tiny{$21$}}$  & $729 \pm \text{\tiny{$6$}}$ \\
 &  & \texttt{jump} & $219 \pm \text{\tiny{$117$}}$ & $175 \pm \text{\tiny{$92$}}$ & $517 \pm \text{\tiny{$113$}}$ & $712 \pm \text{\tiny{$20$}}$ & $737 \pm \text{\tiny{$21$}}$ \\
 &  & \texttt{run} & $97 \pm \text{\tiny{$95$}}$ & $63 \pm \text{\tiny{$151$}}$ & $402 \pm \text{\tiny{$83$}}$ & $512 \pm \text{\tiny{$31$}}$ & $504 \pm \text{\tiny{$13$}}$ \\
 &  & \texttt{stand} & $223 \pm \text{\tiny{$158$}}$ & $65 \pm \text{\tiny{$130$}}$ & $757 \pm \text{\tiny{$178$}}$ & $899 \pm \text{\tiny{$31$}}$ & $955 \pm \text{\tiny{$36$}}$ \\
 &  & \texttt{walk} & $40 \pm \text{\tiny{$109$}}$ & $35 \pm \text{\tiny{$65$}}$ & $385 \pm \text{\tiny{$97$}}$ & $721 \pm \text{\tiny{$43$}}$ & $749 \pm \text{\tiny{$57$}}$ \\
\addlinespace\cline{1-8}\addlinespace
 \multirow[c]{10}{*}{\texttt{\texttt{Walker}}} & \multirow[c]{5}{*}{\texttt{flickering}} & \texttt{overall} & $76 \pm \text{\tiny{$32$}}$ & $82 \pm \text{\tiny{$10$}}$ & $519 \pm \text{\tiny{$37$}}$ & $511 \pm \text{\tiny{$85$}}$ & $637 \pm \text{\tiny{$60$}}$ \\
 &  & \texttt{flip} & $54 \pm \text{\tiny{$11$}}$ & $57 \pm \text{\tiny{$21$}}$ & $450 \pm \text{\tiny{$60$}}$ & $400 \pm \text{\tiny{$79$}}$ & $560 \pm \text{\tiny{$35$}}$ \\
 &  & \texttt{run} & $27 \pm \text{\tiny{$11$}}$ & $34 \pm \text{\tiny{$6$}}$ & $250 \pm \text{\tiny{$22$}}$ & $237 \pm \text{\tiny{$34$}}$ & $359 \pm \text{\tiny{$66$}}$ \\
 &  & \texttt{stand} & $189 \pm \text{\tiny{$82$}}$ & $204 \pm \text{\tiny{$35$}}$ & $712 \pm \text{\tiny{$123$}}$ & $761 \pm \text{\tiny{$77$}}$ & $871 \pm \text{\tiny{$43$}}$ \\
 &  & \texttt{walk} & $32 \pm \text{\tiny{$29$}}$ & $52 \pm \text{\tiny{$9$}}$ & $693 \pm \text{\tiny{$52$}}$ & $646 \pm \text{\tiny{$204$}}$ & $772 \pm \text{\tiny{$138$}}$ \\
\addlinespace\cline{2-8}\addlinespace
 & \multirow[c]{5}{*}{\texttt{noisy}} & \texttt{overall} & $339 \pm \text{\tiny{$47$}}$ & $309 \pm \text{\tiny{$78$}}$ & $427 \pm \text{\tiny{$69$}}$ & $434 \pm \text{\tiny{$23$}}$ & $637 \pm \text{\tiny{$60$}}$ \\
 &  & \texttt{flip} & $220 \pm \text{\tiny{$56$}}$ & $165 \pm \text{\tiny{$72$}}$ & $340 \pm \text{\tiny{$69$}}$ & $361 \pm \text{\tiny{$45$}}$ & $560 \pm \text{\tiny{$35$}}$ \\
 &  & \texttt{run} & $193 \pm \text{\tiny{$49$}}$ & $143 \pm \text{\tiny{$56$}}$ & $165 \pm \text{\tiny{$44$}}$ & $183 \pm \text{\tiny{$17$}}$ & $359 \pm \text{\tiny{$66$}}$ \\
 &  & \texttt{stand} & $527 \pm \text{\tiny{$157$}}$ & $509 \pm \text{\tiny{$137$}}$ & $608 \pm \text{\tiny{$73$}}$ & $731 \pm \text{\tiny{$85$}}$ & $871 \pm \text{\tiny{$43$}}$ \\
 &  & \texttt{walk} & $335 \pm \text{\tiny{$78$}}$ & $387 \pm \text{\tiny{$96$}}$ & $577 \pm \text{\tiny{$112$}}$ & $486 \pm \text{\tiny{$42$}}$ & $772 \pm \text{\tiny{$138$}}$ \\
\bottomrule
\end{tabular}
}
\end{table}

\begin{table}
\centering
\caption{\textbf{Full results on ExORL changed dynamics experiments (5 seeds).} For each dataset-domain pair, we report the score at the step for which the all-task IQM is maximised when averaging across 5 seeds $\pm$ the standard deviation.}
\label{table: full changed dynamics} 
\scalebox{0.80}{
\begin{tabular}{lllllll|l} 
\toprule
 \texttt{Dynamics} & \texttt{Environment} & \texttt{Task} & \texttt{FB} & \texttt{HILP} & \texttt{FB-stack} & \texttt{FB-M (ours)} & \texttt{MDP} \\
\midrule \addlinespace
\multirow[c]{15}{*}{\texttt{Interpolation}} & \multirow[c]{5}{*}{\texttt{Cheetah}} & \texttt{overall} & $476 \pm \text{\tiny{$77$}}$ & $67 \pm \text{\tiny{$37$}}$ & $156 \pm \text{\tiny{$55$}}$ & $453 \pm \text{\tiny{$120$}}$ & $474 \pm \text{\tiny{$129$}}$ \\
 &  & \texttt{run} & $167 \pm \text{\tiny{$59$}}$ & $17 \pm \text{\tiny{$11$}}$ & $59 \pm \text{\tiny{$18$}}$ & $150 \pm \text{\tiny{$68$}}$ & $183 \pm \text{\tiny{$59$}}$ \\
 &  & \texttt{run backward} & $166 \pm \text{\tiny{$21$}}$ & $6 \pm \text{\tiny{$21$}}$ & $36 \pm \text{\tiny{$38$}}$ & $192 \pm \text{\tiny{$66$}}$ & $176 \pm \text{\tiny{$53$}}$ \\
 &  & \texttt{walk} & $816 \pm \text{\tiny{$280$}}$ & $84 \pm \text{\tiny{$43$}}$ & $312 \pm \text{\tiny{$52$}}$ & $483 \pm \text{\tiny{$242$}}$ & $726 \pm \text{\tiny{$194$}}$ \\
 &  & \texttt{walk backward} & $777 \pm \text{\tiny{$71$}}$ & $160 \pm \text{\tiny{$83$}}$ & $186 \pm \text{\tiny{$226$}}$ & $956 \pm \text{\tiny{$167$}}$ & $812 \pm \text{\tiny{$217$}}$ \\
\addlinespace\cline{2-8}\addlinespace
 & \multirow[c]{5}{*}{\texttt{Quadruped}}  & \texttt{overall} & $551 \pm \text{\tiny{$82$}}$ & $186 \pm \text{\tiny{$55$}}$ & $394 \pm \text{\tiny{$76$}}$ & $767 \pm \text{\tiny{$24$}}$ & $729 \pm \text{\tiny{$6$}}$ \\
 &  & \texttt{jump} & $566 \pm \text{\tiny{$128$}}$ & $291 \pm \text{\tiny{$188$}}$ & $412 \pm \text{\tiny{$69$}}$ & $787 \pm \text{\tiny{$22$}}$ & $737 \pm \text{\tiny{$21$}}$ \\
 &  & \texttt{run} & $360 \pm \text{\tiny{$120$}}$ & $51 \pm \text{\tiny{$27$}}$ & $251 \pm \text{\tiny{$54$}}$ & $496 \pm \text{\tiny{$17$}}$ & $504 \pm \text{\tiny{$13$}}$ \\
 &  & \texttt{stand} & $842 \pm \text{\tiny{$79$}}$ & $171 \pm \text{\tiny{$186$}}$ & $521 \pm \text{\tiny{$82$}}$ & $964 \pm \text{\tiny{$9$}}$ & $955 \pm \text{\tiny{$36$}}$ \\
 &  & \texttt{walk} & $434 \pm \text{\tiny{$12$}}$ & $81 \pm \text{\tiny{$68$}}$ & $358 \pm \text{\tiny{$111$}}$ & $803 \pm \text{\tiny{$84$}}$ & $749 \pm \text{\tiny{$57$}}$ \\
\addlinespace\cline{2-8}\addlinespace
 & \multirow[c]{5}{*}{\texttt{Walker}} & \texttt{overall} & $637 \pm \text{\tiny{$41$}}$ & $391 \pm \text{\tiny{$107$}}$ & $603 \pm \text{\tiny{$8$}}$ & $635 \pm \text{\tiny{$19$}}$ & $637 \pm \text{\tiny{$60$}}$ \\
 &  & \texttt{flip} & $452 \pm \text{\tiny{$165$}}$ & $340 \pm \text{\tiny{$89$}}$ & $459 \pm \text{\tiny{$15$}}$ & $452 \pm \text{\tiny{$44$}}$ & $560 \pm \text{\tiny{$35$}}$ \\
 &  & \texttt{run} & $362 \pm \text{\tiny{$33$}}$ & $161 \pm \text{\tiny{$47$}}$ & $236 \pm \text{\tiny{$23$}}$ & $298 \pm \text{\tiny{$16$}}$ & $359 \pm \text{\tiny{$66$}}$ \\
 &  & \texttt{stand} & $887 \pm \text{\tiny{$18$}}$ & $752 \pm \text{\tiny{$290$}}$ & $856 \pm \text{\tiny{$4$}}$ & $890 \pm \text{\tiny{$30$}}$ & $871 \pm \text{\tiny{$43$}}$ \\
 &  & \texttt{walk} & $845 \pm \text{\tiny{$29$}}$ & $316 \pm \text{\tiny{$139$}}$ & $853 \pm \text{\tiny{$28$}}$ & $886 \pm \text{\tiny{$40$}}$ & $772 \pm \text{\tiny{$138$}}$ \\

\addlinespace\cline{1-8} \cline{2-8}\addlinespace
\multirow[c]{15}{*}{\texttt{Extrapolation}} & \multirow[c]{5}{*}{\texttt{Cheetah}} & \texttt{overall} & $369 \pm \text{\tiny{$140$}}$ & $62 \pm \text{\tiny{$33$}}$ & $178 \pm \text{\tiny{$83$}}$ & $586 \pm \text{\tiny{$144$}}$ & $516 \pm \text{\tiny{$23$}}$ \\
 &  & \texttt{run} & $146 \pm \text{\tiny{$92$}}$ & $16 \pm \text{\tiny{$12$}}$ & $18 \pm \text{\tiny{$23$}}$ & $223 \pm \text{\tiny{$73$}}$ & $252 \pm \text{\tiny{$65$}}$ \\
 &  & \texttt{run backward} & $225 \pm \text{\tiny{$83$}}$ & $1 \pm \text{\tiny{$0$}}$ & $70 \pm \text{\tiny{$75$}}$ & $320 \pm \text{\tiny{$128$}}$ & $157 \pm \text{\tiny{$31$}}$ \\
 &  & \texttt{walk} & $366 \pm \text{\tiny{$400$}}$ & $86 \pm \text{\tiny{$90$}}$ & $59 \pm \text{\tiny{$45$}}$ & $814 \pm \text{\tiny{$121$}}$ & $819 \pm \text{\tiny{$78$}}$ \\
 &  & \texttt{walk backward} & $743 \pm \text{\tiny{$230$}}$ & $144 \pm \text{\tiny{$50$}}$ & $312 \pm \text{\tiny{$275$}}$ & $976 \pm \text{\tiny{$292$}}$ & $632 \pm \text{\tiny{$206$}}$ \\
\addlinespace\cline{2-8}\addlinespace
 & \multirow[c]{5}{*}{\texttt{Quadruped}} & \texttt{overall} & $333 \pm \text{\tiny{$61$}}$ & $120 \pm \text{\tiny{$47$}}$ & $263 \pm \text{\tiny{$47$}}$ & $704 \pm \text{\tiny{$31$}}$ & $645 \pm \text{\tiny{$52$}}$ \\
 &  & \texttt{jump} & $309 \pm \text{\tiny{$46$}}$ & $131 \pm \text{\tiny{$81$}}$ & $272 \pm \text{\tiny{$43$}}$ & $714 \pm \text{\tiny{$79$}}$ & $615 \pm \text{\tiny{$81$}}$ \\
 &  & \texttt{run} & $212 \pm \text{\tiny{$42$}}$ & $42 \pm \text{\tiny{$41$}}$ & $170 \pm \text{\tiny{$33$}}$ & $474 \pm \text{\tiny{$7$}}$ & $360 \pm \text{\tiny{$29$}}$ \\
 &  & \texttt{stand} & $510 \pm \text{\tiny{$121$}}$ & $275 \pm \text{\tiny{$191$}}$ & $334 \pm \text{\tiny{$39$}}$ & $957 \pm \text{\tiny{$22$}}$ & $716 \pm \text{\tiny{$117$}}$ \\
 &  & \texttt{walk} & $268 \pm \text{\tiny{$60$}}$ & $62 \pm \text{\tiny{$52$}}$ & $274 \pm \text{\tiny{$78$}}$ & $723 \pm \text{\tiny{$136$}}$ & $420 \pm \text{\tiny{$50$}}$ \\
\addlinespace\cline{2-8}\addlinespace
 & \multirow[c]{5}{*}{\texttt{Walker}} & \texttt{overall} & $316 \pm \text{\tiny{$80$}}$ & $146 \pm \text{\tiny{$74$}}$ & $463 \pm \text{\tiny{$15$}}$ & $478 \pm \text{\tiny{$19$}}$ & $555 \pm \text{\tiny{$89$}}$ \\
 &  & \texttt{flip} & $86 \pm \text{\tiny{$18$}}$ & $107 \pm \text{\tiny{$29$}}$ & $320 \pm \text{\tiny{$10$}}$ & $336 \pm \text{\tiny{$86$}}$ & $370 \pm \text{\tiny{$48$}}$ \\
 &  & \texttt{run} & $218 \pm \text{\tiny{$41$}}$ & $81 \pm \text{\tiny{$31$}}$ & $283 \pm \text{\tiny{$19$}}$ & $297 \pm \text{\tiny{$42$}}$ & $301 \pm \text{\tiny{$74$}}$ \\
 &  & \texttt{stand} & $475 \pm \text{\tiny{$261$}}$ & $290 \pm \text{\tiny{$190$}}$ & $624 \pm \text{\tiny{$34$}}$ & $691 \pm \text{\tiny{$64$}}$ & $715 \pm \text{\tiny{$138$}}$ \\
 &  & \texttt{walk} & $501 \pm \text{\tiny{$55$}}$ & $98 \pm \text{\tiny{$74$}}$ & $632 \pm \text{\tiny{$57$}}$ & $574 \pm \text{\tiny{$77$}}$ & $476 \pm \text{\tiny{$137$}}$ \\
\bottomrule
\end{tabular}
}
\end{table}

\end{document}